\documentclass{elsart}

\usepackage{graphicx}

\usepackage{amssymb}

\begin{document}

\begin{frontmatter}



\title{Neural Network Ensembles: Evaluation of Aggregation Algorithms}


\author{P.M. Granitto, P.F. Verdes and H.A. Ceccatto}

\address{Instituto de F\'{\i}sica Rosario, CONICET/UNR,
Boulevard 27 de Febrero 210 Bis, 2000 Rosario, Rep\'ublica
Argentina}

\begin{abstract}
Ensembles of artificial neural networks show improved
generalization capabilities that outperform those of single
networks. However, for aggregation to be effective, the individual
networks must be as accurate and diverse as possible. An important
problem is, then, how to tune the aggregate members in order to
have an optimal compromise between these two conflicting
conditions. We present here an extensive evaluation of several
algorithms for ensemble construction, including new proposals and
comparing them with standard methods in the literature. We also
discuss a potential problem with sequential aggregation
algorithms: the non-frequent but damaging selection through their
heuristics of particularly bad ensemble members. We introduce
modified algorithms that cope with this problem by allowing
individual weighting of aggregate members. Our algorithms and
their weighted modifications are favorably tested against other
methods in the literature, producing a sensible improvement in
performance on most of the standard statistical databases used as
benchmarks.
\end{abstract}


\end{frontmatter}


\section{Introduction} \label{}
For most regression and classification problems, combining the
outputs of several predictors improves on the performance of a
single generic one \cite{Sharkey}. Formal support to this property
is provided by the so-called bias/variance dilemma \cite{Geman},
based on a suitable decomposition of the prediction error.
According to these ideas, good ensemble members must be both {\it
accurate} and {\it diverse}, which poses the problem of generating
a set of predictors with reasonably good individual performances
and independently distributed predictions for the test points.

Diverse individual predictors can be obtained in several ways.
These include: i) using different algorithms to learn from the
data (classification and regression trees, artificial neural
networks, support vector machines, etc.), ii) changing the
internal structure of a given algorithm (for instance, number of
nodes/depth in trees or architecture in neural networks), and iii)
learning from different adequately-chosen subsets of the data set.

The probability of success in strategy iii), the most frequently
used, is directly tied to the instability of the learning
algorithm \cite{Breiman}. That is, the method must be very
sensitive to small changes in the structure of the data and/or in
the parameters defining the learning process. Again, classical
examples in this sense are classification and regression trees and
artificial neural networks (ANNs). In particular, in the case of
ANNs the instability comes naturally from the inherent data and
training process randomness, and also from the intrinsic
non-identifiability of the model.

The combination of strong instability of the learning algorithm
with the trade-off predictors' diversity vs. good individual
generalization capabilities requires an adequate selection of the
ensemble members. Attempts to achieve a good compromise between
the above mentioned properties include elaborations of two general
techniques: {\it bagging} \cite{Breiman} and {\it boosting}
\cite{Freund}. These standard methods for ensemble construction
follow two different strategies: Bagging (short for 'bootstrap
aggregation'), and variants thereof, train independent predictors
on bootstrap re-samples ${\emph L}_{n}$ ($n=1,M$) of the available
data ${\emph D}$, usually employing the unused examples ${\emph
V}_{n}={\emph D}-{\emph L}_{n}$ for validation purposes. These
predictors are then aggregated according to different rules (for
instance, simple or weighted average). Boosting and its variants
are stagewise procedures that, starting from a predictor trained
on ${\emph D}$, sequentially train new aggregate members on
bootstrap re-samples drawn with modified probabilities. According
to the general approach, each example in ${\emph D}$ is given a
different chance to appear in a new training set by prioritizing
patterns poorly learnt on previous stages. In the end, the
predictions of the different members so generated are weighted
with a decreasing function of the error each predictor makes on
its training data.

For regression problems, on which we will focus here, boosting is
still a construction area, where no algorithm has emerged yet as
'the' proper way of implementing this technique
\cite{Freund,Druckerb,Friedman,Avnimelech,Cara,Zemel}.
Consequently, bagging is the most common method for ANN
aggregation. On the other hand, intermediate alternatives between
bagging and boosting, which optimize directly the ensemble
generalization performance instead of seeking for the best
individual members, have not been much explored \cite{Rosen}. In
this work we compare different strategies for ensemble
construction, restricting ourselves to work in the regression
setting and using ANNs as learning method. These restrictions are
not essential; in principle, our analysis can be extended to
classification problems and to other regression/classification
methods. Furthermore, we will discuss stepwise algorithms to build
the best aggregate after network training, thus incorporating the
condition of optimal ensemble performance. Our main purpose is to
establish rules as general as possible to build accurate
regression aggregates. For this, we will

\begin{itemize}
\item {discuss, in a unifying picture, several alternatives
already proposed in the literature for the aggregation of ANNs,}
\item{present a new algorithm that is optimal within this
unified point of view,}
\item{propose a simple weighting scheme of ensemble members that
improves the aggregates' generalization performances, and}
\item{perform an extensive comparison of all these methods among
themselves and with boosting techniques on several synthetic and
real-world data sets.}
\end{itemize}

The organization of this work is the following: In Section 2 we
re-discuss several bagging-like methods proposed in the
literature, considering them as different strategies for selecting
the termination point of training processes for ensemble members.
In this section we also present a new algorithm that is optimal
from this point of view. In Section 3 we introduce the synthetic
and real-world databases considered in this study, and describe
the experimental settings used to learn from them. In Section 4 we
obtain empirical evidence on the relative efficacy of all the
methods discussed in Section 2 by applying them to these
databases. Then, in Section 5 we present a modified, weighted
version of the best algorithms and test their performances by
comparison with the results in Section 4 and also against boosting
and other techniques. Finally, in Section 6 we summarize the work
done and the main results obtained, and draw some conclusions.

\section{Ensemble Construction Algorithms}

The simplest way of generating a regressor aggregate is bagging
\cite{Breiman}. According to this method, from the data set
${\emph D}$ containing $N$ examples ($t,\mathbf{x}$) one generates
bootstrap re-samples ${\emph L}_n$ ($n=1,M$) by drawing with
replacement $N$ training patterns. Thus, each training set ${\emph
L}_n$ will contain, on average, $0.63N$ different examples, some
of them repeated one or more times \cite{Efron}. The remaining
$0.37N$ examples in ${\emph V}_n = {\emph D}-{\emph L}_n$ are
generally used for validation purposes in the regressor learning
phase (backpropagation training of the ANN in our case). In this
way one generates $M$ different members $f_{n}$ of the ensemble,
whose outputs on a test point $\mathbf x$ are finally averaged to
produce the aggregate prediction $\Phi({\mathbf x})= w_1
f_1({\mathbf x}) + ... + w_M f_M ({\mathbf x})$. The weights $w_n$
are usually taken equal to $1/M$ (simple averaging). Other options
will be discussed in the next section. Notice that, according to
this method, all the regressors are trained independently and
their performances individually optimized using the ``out-of-bag"
data in ${\emph V}_{n}$. Then, although there is no fine-tuning of
the ensemble members' diversity, the method frequently improves
largely on the average performance of the single regressors
$f_{n}$.

Bagging can be viewed as a first stage in a sequence of
increasingly more
sophisticated algorithms for building a
composite ANN regressor. To understand this, let's consider first
the situation in which a common validation subset ${\emph V}$ of
the dataset ${\emph D}$ is kept unseen by all the networks during
their training phases. Let's also consider training to convergence
$M$ ANNs on bootstrap re-samples ${\emph L}_n$ obtained now from
${\emph L}={\emph D}-{\emph V}$, saving the intermediate states
$f_{n}(\tau)$ at each training epoch $\tau$ ({\it i.e.},
$f_{n}(\tau)$ is the ANN model whose weights and biases take the
values obtained at epoch $\tau$ of the training process). Building
an ensemble is then translated to the task of selecting a
combination of one state $f_n(\tau_n^{\mathrm {opt}})$ from each
of the $M$ runs to create an optimal ensemble, that is, an
ensemble with the smallest error on ${\emph V}$. In this light,
bagging solves the problem by choosing the state using only
information on the given run ($\tau_n^{\mathrm {opt}}$ is the
number of training epochs for which the validation error on
${\emph V}$ is minimum). In more advanced algorithms, the
regressors are not optimized individually but as part of the
aggregate. For ANNs, the simplest way of doing this is choosing a
(common) optimal number of training epochs $\tau_n^{\mathrm
{opt}}=\tau^{\mathrm {opt}}$ for all networks by optimizing the
{\it ensemble} performance on ${\emph V}$ \cite{Naftaly}:

\begin{equation}\label{tau}
\tau^{\mathrm {opt}} = {\mathrm {argmin}}_{\tau} \sum_{(t,{\mathbf
x}) \in {\emph V}}[t - \Phi({\mathbf x},{\mathbf w}(\tau))]^2 .
\nonumber
\end{equation}
Here ${\mathbf w}(\tau)$ are the ANN internal parameters (weights
and biases) at epoch $\tau$. Thus, instead of validating the
ensemble members one by one to maximize their individual
performances as in bagging, the algorithm selects a common optimal
stopping point $\tau^{\mathrm{opt}}$ for all the networks in the
ensemble. In practice, one finds that $\tau^{\mathrm{opt}}$ is in
general larger than the individual stopping points found in
bagging, {\it i.e.}, some controlled degree of single network
overfitting improves the aggregate's performance. In the following
we will refer to this algorithm as ``Epoch".

The above described strategy can be further pushed on by selecting
not a single optimal $\tau^{\mathrm{opt}}$ for all networks but
independent $\tau^{\mathrm{opt}}_{n}$ for each network in the
ensemble. This requires minimizing

\begin{equation}\label{Esa}
 E({\vec{\tau}}) = \sum_{(t,{\mathbf x}) \in {\emph V}} [t - \Phi
 ({\mathbf x},{\mathbf w}({\vec{\tau}}))]^2 \nonumber
\end{equation}
as a function of the set of training epochs
${\vec{\tau}}=\{\tau_{n}; n=1,M\}$ for all networks. This can be
accomplished, for instance, by using simulated annealing in
${\vec{\tau}}$-space. That is, starting from networks trained
${\vec{\tau}}_0$ epochs, we randomly change $\tau_{0n}$ and check
whether the ensemble generalization error (\ref{Esa}) increases or
decreases when network $n$ is trained up to $\tau_{0n}+\Delta
\tau$. As usual, we accept the move with probability 1 when
$E({\vec{\tau}})$ decreases, and with probability
\begin{equation}\label{}
{\frac {\exp\{-\beta [E({\vec{\tau}})-E({\vec{\tau}}_0)]\}} {1+
\exp\{-\beta [E({\vec{\tau}})-E({\vec{\tau}}_0)]\}} } \nonumber
\end{equation}
when $E({\vec{\tau}})$ increases. This is repeated many times
considering different networks $n$ (chosen either at random or
sequentially), while the annealing parameter $\beta$ is
conveniently increased at each step; the algorithm runs until
$E({\vec{\tau}})$ settles in a deep local minimum. In practice we
have taken $\Delta \tau = r \tau^{\mathrm {max}}/20$, where
$\tau^{\mathrm {max}}$ is the maximum number of training epochs
and $r$ is a random number in the interval $[-1,1]$. The annealing
temperature was decreased according to $\beta^{-1}=0.995^{q}
E({\vec {\tau_0}})/2$, where $q$ is the annealing step. We point
out that the minimization problem is simple enough not to depend
critically on these choices. As far as we know, this algorithm
---which we will call ``SimAnn"--- has not been previously
discussed in the literature and constitutes one of the main
contributions of this work. Notice that for its implementation, as
well as for the simplest implementation of Epoch, one is forced to
store all the intermediate networks $f_{n} [{\mathbf w}(\tau)]$.
However, given the large storage capacity in computers nowadays,
in most applications this requirement is not severe.

In the common situation of scarcity of data, the need to keep an
independent validation set ${\emph V}$ is a serious drawback that
limits the efficacy of the methods discussed above. An alternative
approach is to resort to the out-of-bag patterns $(t_p,{\mathbf
x}_p)\in {\emph V}_{n}$ unseen by network $f_{n}$, and optimize
with respect to the number of training epochs the error

\begin{equation}\label{Esa1}
 E({\vec{\tau}}) = \sum_{p=1}^{N} [t_p - \Phi_{p}
 ({\mathbf x}_p,{\mathbf w}({\vec{\tau}}))]^2 .
\end{equation}
Here $\Phi_p ({\mathbf x}_p,{\mathbf w}(\vec {\tau})) =
\sum_{n=1,M} w_{pn}f_{n}[{\mathbf x}_p,{\mathbf w}(\vec {\tau})]$
is the aggregate regressor built with those networks that have not
seen pattern $(t_p,{\mathbf x}_p)$ in their training phase, {\it
i.e.}
\begin{equation}\label{}
w_{pn}= \frac {\gamma_{pn}} {\sum_n \gamma_{pn}} , \nonumber
\end{equation}
where $\gamma_{pn}=1$ if $(t_p,{\mathbf x}_p)\in {\emph V}_{n}$
and 0 otherwise. Notice that the validation procedure generated by
Eq. (\ref{Esa1}) amounts to effectively optimizing the
performances of several subsets of the $M$ trained ANNs, each
subset including on average $0.37M$ networks. The advantage is
that, like in the description of bagging at the beginning of this
section, no sub-utilization of data for validation purposes is
necessary.

The above described strategy can be slightly simplified by
selecting independent $\tau^{\mathrm{opt}}_n$ for each network in
the ensemble. This is the proposal of the so-called NeuralBAG
algorithm \cite{Carney}, which chooses

\begin{equation}\label{nb}
\tau^{\mathrm{opt}}_n = {\mathrm {argmin}}_{\tau}
\sum_{(t_p,{\mathbf x}_p) \in {\emph V}_n} [t_p -
\Phi_{p}({\mathbf x}_p,{\mathbf w}(\tau))]^2
\end{equation}
This is a rather {\it ad hoc} criterion: notice that in (\ref{nb})
the networks $f_m$ with $m \neq n$ are trained up to
$\tau^{\mathrm {opt}}_n$, but they are effectively trained
$\tau^{\mathrm{opt}}_m$ epochs in the final ensemble.
Nevertheless, judging from the reported results \cite{Carney}, it
seems to be effective in practice.

All the strategies for ANN aggregation discussed so far minimize
some particular error function in a global way. A different
approach is to adapt the typical hill-climbing search method to
this problem. In a previous work \cite{Granitto} we proposed a
simple way of generating a ANN ensemble through the sequential
aggregation of individual predictors, where the learning process
of a new ensemble member is validated by the previous-stage
aggregate prediction performance. That is, the early-stopping
method is applied by monitoring the generalization capability on
${\emph V}_{\mathrm {n+1}}$ of the $n$-stage aggregate predictor
plus the $n+1$ network being currently trained. In this way we
retain the simplicity of independent network training and only the
validation process becomes slightly more involved, leading again
to a controlled overtraining (``late-stopping") of the individual
networks. Notice that, despite the stepwise characteristic of this
algorithm (here called SECA, for Stepwise Ensemble Construction
Algorithm), it can be implemented after the parallel training of
networks if desirable. Alternatively, if implemented sequentially
it avoids completely the burden of storing networks at
intermediate training times like in the algorithms described
above.

For the sake of completeness, we summarize the implementation of
SECA as follows:

{\textbf{Step 1}: Generate a training set ${\emph L}_1$ by a
bootstrap re-sample from dataset ${\emph D}$, and a validation set
${\emph V}_1 = {\emph D}-{\emph L}_1$ by collecting all instances
in ${\emph D}$ that are not included in ${\emph L}_1$. Produce a
model $f_1$ by training a network on ${\emph L}_1$ until a minimum
$e_f({\emph V}_1)$ of the generalization error on ${\emph V}_1$ is
reached.

{\textbf{Step 2}: Generate new training and validation sets
${\emph L}_2$ and ${\emph V}_2$ respectively, using the procedure
described in Step 1. Produce a model $f_2$ training a network
until the generalization error on ${\emph V}_2$ of the aggregate
predictor $\Phi_2 =(f_1+f_2)/2$ reaches a minimum $e_\Phi({\emph
V}_2)$. In this step the parameters of model $f_1$ are kept
constant and the model $f_2$ is trained with the usual (quadratic)
cost function on ${\emph L}_2$.

{\textbf{Step 3}: Iterate the process until a number $M$ of models
is produced. A suitable $M$ can be estimated from the behavior of
$e_\Phi({\emph V}_n)$ as a function of $n$, since this error will
stabilize when adding more networks to the aggregate becomes
useless.

In this algorithm the individual networks are directly trained
with a late-stopping method based on the current ensemble
generalization performance. The method seems to reduce the
aggregate generalization error without paying much attention to
whether this improvement is related to enhancing the members'
diversity or not. However, one can see \cite{Granitto} that it
actually finds diverse models to reduce the ensemble error by
looking, at every stage, for a new model anticorrelated with the
current ensemble. Notice that SECA can be also implemented using
an external validation set $\emph V$, in which case all the
bootstrap complements ${\emph V}_n$ are replaced by this fixed
set.

All the above described methods constitute a chain of increasingly
optimized algorithms for ensemble building, starting from the
simplest Bagging idea of optimizing networks independently to
SimAnn, which should produce the ``optimal" ensemble ({\it i.e.},
the ensemble with the minimum validation error \ref{Esa1}). Let's
consider a simple analysis of the computational cost involved in
the implementation of these algorithms. Once the $M$ ANNs have
been independently trained and $T$ networks saved along each
training evolution, which is common to all the algorithms, Bagging
requires a computational time $t \sim M \times T$ to select the
best combination (essentially, the evaluation of the $T$ ANN's
validation errors for each of the $M$ networks to find the
corresponding minima). Epoch requires exactly the same
computational effort to find the (common) optimal stopping point
for all networks. NeuralBag uses, instead, $t \sim M^2 \times T$
evaluations to find the best aggregate. Finally, SECA and SimAnn
require ${\frac {(M+1)} {2}}\times M \times T$ and $p \times M
\times T$ network evaluations, respectively. Here we have written
the number of simulated annealing steps $N_{\mathrm {sa}} = pT$,
with $p$ an arbitrary integer, to facilitate the comparison. In
the following we will take $p \sim M$ to have a fair comparison
between NeuralBag, SECA and SimAnn. Notice, however, that the
major demand from a computational point of view is the ANN
training and not the network selection to build the ensemble. In
practice, in the algorithms' evaluations in Section 4 and 5 we
have taken $M=20$, $T=200$ and $p=15$, with all the networks
trained a maximum of $10T$ to $100T$ epochs, depending on the
database.

As mentioned in the Introduction, a completely different strategy
for building composite regression/classification machines is
boosting. For classification problems, its main difference with
bagging is the use of modified probabilities to re-sample the
training sets ${\emph L}_n$.  At stage $n$, the weights associated
to examples in $\emph D$ are larger for those examples poorly
learnt in previous stages, so that they eventually appear several
times in ${\emph L}_n$. In this way, the new predictor $f_n$
trained on ${\emph L}_n$ specializes on these hard examples.
Finally, the inclusion of $f_n$ in the ensemble with a
suitably-chosen weight allows the exponential decrease with
boosting rounds $n$ of the ensemble's training error on the whole
dataset $\emph D$. Notice that, in addition to the above mentioned
modification of re-sampling probabilities, other differences with
bagging are: i) boosting is essentially a stage-wise approach,
which requires a sequential training of the aggregate members
$f_n$, and ii) in the final ensemble these members are weighted
according to their performances on the respective training sets
${\emph L}_n$ (using a decreasing function of the training error).
A further consideration of this last characteristic will be done
in Section 4, where we discuss a weighting scheme for bagged
regressors alternative to the simple average considered in this
section.

While boosting is, as explained above, a well defined procedure in
the classification setting, for regression problems there are
several ways of implementing its basic ideas. Unfortunately, none
of them has yet emerged as ``the" proper way of boosting
regressors. Without the intention of exhausting all the proposed
implementations, we can distinguish two boosting strategies for
solving regression problems: i) by forward stage-wise additive
modelling, which modifies the target values to effectively fit
residual errors\cite{Friedman,Cara,Duffy}, and ii) by reducing the
regression problem to classification and essentially changing
example weights to emphasize those which were poorly learnt on
previous stages of the fitting process
\cite{Freund,Ridgeway,Druckerb,Zemel}. In order to compare with
the bagging-like algorithms described above, in this work we will
implement the boosting techniques from \cite{Friedman} and
\cite{Druckerb} as examples of these two different strategies.

In Sections 4 and 5 we will show how all the heuristic algorithms
described in this section work on real and synthetic data. This
will provide a fairly extensive comparison of the already known
methods and will test the new SimAnn algorithm against all the
other methods. In the next section we briefly describe the
databases and experimental settings considered for this
comparison.

\section{Benchmark Databases and Experimental Settings}

We have evaluated the algorithms described in the previous section
by applying them to several benchmark databases: the synthetic
Friedman \#1, 2, 3 data sets and chaotic Ikeda map, and the
real-world Abalone, Boston Housing, Ozone and Servo data sets. In
the cases of the Friedman data sets we can control the (additive)
noise level, which allows us to investigate its influence on the
different algorithm's performances. We present the results for the
Ikeda map together with those of real-world sets because the level
of noise in this problem is fixed by its intrinsic dynamics. In
addition, at the end of next section we will present results on
the Mackey-Glass equation, which allows a more general comparison
with other regression methods in the literature previously applied
to this problem \cite{Muller,Ratsch}.

In the following we give brief descriptions of the databases and
the ANN architectures used. In all cases, the number of hidden
units $h$ have been selected by trial and error, using a
validation set and looking for the minimum generalization error on
this set as a function of $h$. Once the network architecture was
chosen, it was kept the same during all the calculations. Notice
that this is not a particularly important point, since we want to
compare the efficacy of different aggregation methods and all of
them use the same trained networks.
\begin{itemize}
\item{ Friedman \#1}

The Friedman \#1 synthetic data set corresponds to training
vectors with 10 input and one output variables generated according
to
\begin{equation}
t = 10\sin(x_1 x_2) + 20(x_3 - 0.5)^2 + 10x_4 + 5x_5 +
\varepsilon,
\end{equation}
where $\varepsilon$ is Gaussian noise and $x_1,\ldots x_{10}$ are
uniformly distributed over the interval $[0,1]$. Notice that
$x_6,\dots x_{10}$ do not enter in the definition of $t$ and are
only included to check the prediction method's ability to ignore
these inputs. In order to explore the algorithm's performances in
different situations we considered different noise levels and
training set lengths. The Gaussian noise component was
alternatively set to: $\varepsilon = 0$ (No noise, labeled
``free"), $\varepsilon$ with normal distribution ${\emph N}(\mu=0,
\sigma=1)$ (low noise), and $\varepsilon$ with normal distribution
${\emph N}(\mu=0, \sigma=2))$ (high noise). We generated 1200
sample vectors for each noise level and these data sets were
randomly split in training and test sets. The training sets
${\emph D}$ had alternatively 50, 100 and 200 patterns, while the
test set contained always 1000 examples. We considered ANNs with
10:$h$:1 architectures, with the number of hidden units $h = 6,
10$ and 15 for increasing number of patterns in the training
set.\\

\item{ Friedman \#2}

Friedman \#2 has four independent variables and the target data
are generated according to
\begin{equation}
y=x_1^2+\sqrt{x_2 x_3 - (x_2 x_4)^{-2}}+ \varepsilon \nonumber
\end{equation}
where the zero-mean, normal noise is adjusted to give
noise-to-signal power ratios of 0 (no noise), 1:9 (low noise) and
1:3 (high noise). The variables $x_i$ are uniformly distributed in
the ranges
\begin{equation}
0 <  x_1 < 100, \ \ \  20 < \frac{x_2}{2\pi} < 280, \ \ \    0 <
x_3 < 1, \ \ \  1 < x_4 < 11 \nonumber
\end{equation}
The training sets contained 20, 50 and 100 patterns, and the test
set had always 1000 patterns. We considered 4:$h$:1 ANNs, with
$h=4,6$ and 8 according to the training set length.\\

\item{Friedman \#3}

Friedman \#3 has also four independent variables distributed as
above but the target data are generated as
\begin{equation}
y=\tan^{-1}\left[{\frac {x_2 x_3 - (x_2 x_4)^{-2}} {x_1} }\right]
+ \varepsilon \nonumber
\end{equation}
The noise-to-signal ratios were chosen as before, but in this case
the training sets contained 100, 200 and 400 patterns.
Accordingly, we considered $h=6, 8$ and 12. As in the previous
cases, the test sets had always 1000 patterns.\\

\item{Abalone}

The age of abalone is determined by cutting the shell through the
cone, staining it, and counting the number of rings through a
microscope. To avoid this boring task, other measurements easier
to obtain are used to predict the age. Here we considered the data
set that can be downloaded from the UCI Machine Learning
Repository (ftp to ics.uci.edu/pub/machine-learning-databases),
containing 8 attributes and 4177 examples without missing values.
Of these, 1045 patterns were used for testing and 3132 for
training (for all real-world problems considered, the data set
splitting in learning and test sets was chosen following
\cite{Breiman1}). The ANNs used to learn from this set had a 8:5:1
architecture.\\

\item{Boston Housing}

This data set consists of 506 training vectors, with 11 input
variables and one target output. The inputs are mainly
socioeconomic information from census tracts on the greater Boston
area and the output is the median housing price in the tract.
These data can also be downloaded from the UCI Machine Learning
Repository.

Here we considered 450 training examples and 56 data points for
the test set. The ANNs used had a 11:5:1 architecture.\\

\item{Ozone}

The Ozone data correspond to meteorological information (humidity,
temperature, etc.) related to the maximum daily ozone (regression
target) at a location in Los Angeles area. Removing missing values
one is left with 330 training vectors, containing 8 inputs and one
target output in each one. The data set can be downloaded by ftp
(to ftp.stat.berkeley.edu/pub/users/breiman) from the Department
of Statistics, University of California at Berkeley.

We considered ANNs with 8:5:1 architectures and performed a
(random) splitting of the data in training and test sets
containing, respectively, 295 and 35 patterns.\\

\item{Servo}

The servo data cover an extremely non-linear phenomenon
--predicting the rise time of a servomechanism in terms of two
(continuous) gain settings and two (discrete) choices of
mechanical linkages. The set contains 167 instances and can be
downloaded from the UCI Machine Learning Repository.

We considered 4:15:1 ANNs, using 150 examples for training and 17
examples for testing purposes.\\

\item{Ikeda}

The Ikeda laser map \cite{Ikeda}, which describes instabilities in
the transmitted light by a ring cavity system, is given by the
real part of the complex iterates
\begin{equation}
z_{n+1} = 1+0.9 z_n \exp \left[0.4i -
{\frac{6i}{(1+|z_n|^2)}}\right] . \nonumber
\end{equation}
Here we have generated 1100 iterates, using 100 in the training
set and 1000 for testing purposes.

After some preliminary investigations, we chose an embedding
dimension 5 for this map and considered ANNs with a 5:10:1
architecture.

\end{itemize}

For each one of these databases we trained $M =20$ independent
networks, storing $T=200$ intermediate weights and biases
${\mathbf w}({\vec{\tau}})$ on long training experiments until
convergence ($10T$ to $100T$ epochs, depending on the database).
We considered this number of networks after checking on
preliminary evaluations that there were no sensible performance
improvements with bigger ensembles. With these 20 ANNs we
implemented the different bagging-like ensemble construction
algorithms, changing the training stopping points of individual
networks according to the criteria discussed in the previous
section. We did this for the following two different validation
scenarios:
\begin{itemize}
\item Keeping an external validation set $\emph V$, randomly selected
from the data set $\emph D$, and training the 20 ANNs on different
bootstrap re-samples of ${\emph L}={\emph D}-{\emph V}$. Here we
considered two partitions of $\emph D$: 20/80\% and 37/63\%
(following the bootstrap proportion), where in each case the first
number indicates the fraction of data points in $\emph V$. For
this case only the bagging-like algorithms discussed in the
previous section were considered.

\item Validating the training process directly with the out-of-bag
data, as explained in the previous section. This procedure makes
full use of the available data, and in general should produce
better results than the previous situation. In this case we tested
bagging-like techniques and also boosted ANNs according to the
Friedman \cite{Friedman} and Drucker \cite{Druckerb} algorithms,
considering a maximum of 20 boosting rounds for comparison.

\end{itemize}

The results given in the following section correspond to an
average over 50 independent runs of the above-described
procedures, without discarding any anomalous case (for Boston,
Ozone and Servo databases we averaged over 100 experiments because
the smaller test sets allow larger sample fluctuations). We will
not indicate the variance of average errors, since these
deviations only characterize the dispersion in performances due to
different realizations of training and test sets. They have no
direct relevance in comparing the average performances of
different methods (in each run all the algorithms use the same 20
networks). This procedure guarantees that differences in the final
ensemble performances are only due to the aggregation methods
and/or validation settings.

Finally, at the end of Section 5 we compare the best performing
algorithms here considered with several other methods in the
literature. For this comparison we use the chaotic Mackey-Glass
time series:

\begin{itemize}

\item{Mackey-Glass}

The Mackey-Glass time-delay differential equation is a model for
blood cell regulation. It is defined by
\begin{equation}
{\frac {dx(t)} {dt}} = {\frac {0.2x(t-\tau)} {1+x^{10}(t-\tau)}} -
0.1x(t)
\end{equation}
When $x(0) = 1.2$ and $\tau = 17$, we have a non-periodic and
non-convergent time series that is very sensitive to initial
conditions (we assume $x(t)=0$ when $t<0$).

In order to compare with the results in \cite{Muller} and
\cite{Ratsch}, we have downloaded the database used by these
authors and considered, like in these works, an embedding
dimension $d=6$ and 1194 patterns for training and 1000 patterns
for testing purposes. For this problem we took $h=40$.
\end{itemize}

\section{Evaluation Results}

The results quoted below are given in terms of the normalized
mean-squared test error:

\begin{equation}
NMSE_{\emph T} = {\frac {MSE_{\emph T}} {\sigma^2_{\emph D}}} ,
\end{equation}
defined as the mean-squared error on the test set $\emph T$
divided by the variance of the total data set $\emph D$. According
to this definition, $NMSE \simeq 1$ for a constant predictor equal
to the data average and 0 for a perfect one. Then, its value
allows to appraise both the predictor's performance and the
relative complexity of the different regression tasks. Notice
that, as indicated in the table captions, the results are given in
units of $10^{-2}$, so that all the errors are much smaller than 1
and, consequently, the predictions much better than the trivial
data average.

In Tables 1a and 1b we present results for synthetic and real
databases respectively, in the situation in which an external
validation set containing 20\% of the data is used. Tables 2a and
2b correspond to the same case but with 37\% of validation data.
As mentioned in the previous section, here only bagging-like
algorithms are compared. Tables 3a and 3b present the
corresponding results for all bagging-like algorithms and
out-of-bag validation (no hold out data).


First, for external validation, the experiments indicate that
Epoch performs better than Bagging in only 2 of the 27 cases
corresponding to synthetic databases (Friedman \#1,2 and 3, Tables
1a and 2a), and in 2 or none out of 5 cases for the real-world
databases (Tables 1b and 2b), depending on the validation set
size. This poor performance becomes slightly better for out-of-bag
validation (9/27, Table 3a, and 3/5, Table 3b, respectively).
Consequently, we do not find any advantage in using this algorithm
instead of Bagging. Something similar happens with NeuralBag,
which, in spite of the good results presented in \cite{Carney}, in
our experiments only improves on Bagging in roughly half the
cases. On the contrary, both SECA and SimAnn are clearly better
than Bagging: on average, both methods outperform Bagging
approximately in 21 of the 27 Friedman problems and in all but one
case for real databases, independently of the validation used.

Considering all the methods together, Table 1a shows that, for the
27 learning problems associated to the Friedman synthetic data, in
22 cases the best method is either SECA or the network selection
via simulated annealing (SimAnn). This pattern is confirmed by the
results in Table 2a, where SECA and SimAnn are again the best
performers in 22 of the 27 cases. For the real-world databases
these two methods outperform the other bagging-like algorithms in
all cases (see Tables 1b and 2b), with a particularly good
performance of SimAnn. For out-of-bag validation, Table 3a shows
that, consistently with the previous results, in 19 of the 27
experimental situations SECA and SimmAnn are the best performers.
We stress, however, the good performance of Epoch on Friedman \#2
data set, particularly for noise-free data. For the real-world
databases, Table 3b shows that SECA and SimAnn produced the best
results in all but one (Servo) of the regression problems
investigated. All these results obey the expected behaviors with
noise level and data set length. Furthermore, for the synthetic
Friedman problems in general the test error is larger when more
data are held out for validation, although this not the case for
the real-world Abalone, Boston and Ozone datasets. Moreover, for
bagged regressors, independently of the method used to ensemble
them, in all cases the out-of-bag validation is more efficient
than keeping an external set, in agreement with other works in the
literature\cite{Breiman2}. Notice, however, that this last
observation is not valid in the noisy Friedman \#2 and Servo
problems for a single ANN.

In the case of out-of-bag validation, we have performed a paired
$t$-test to check whether SECA and SimAnn significantly outperform
Bagging. Following the procedure in \cite{Druckera}, we considered
a binary variable that assumes a value of 1 when SECA/SimAnn is
better than Bagging and 0 otherwise. If the average of this
variable differs from 0.5 and this difference is statistically
significant, we can affirm that one of the methods is better than
the other. The results of the $t$-test are given in Tables 4a and
4b. For Friedman \#1, SECA and SimAnn are better than Bagging in
all cases; for Friedman \#2 there are no clear differences between
the methods, and for Friedman \#3 SECA and SimAnn are
significantly better than Bagging except for high noise and few
patterns in the training set. For the real-world databases, SECA
and SimAnn are always better than Bagging, with more than 95\% of
statistical significance in several cases. These results are in
complete agreement with the NMSE comparison in Tables 1-3.


\subsection{Accuracy vs. Diversity}
In order to gain some insight into SECA and SimAnn's behaviors
that might explain the good performances shown in Tables 1-4, we
have investigated the standard bias-variance decomposition of the
generalization error\cite{Geman}.

Consider general regression problems where vectors
$\mathbf{x}$ of predictor variables are obtained from some
distribution $P(\mathbf{x)}$ and regression targets $t$ are
generated according to $t=f(\mathbf{x})+\varepsilon$. Here $f$ is
the true regression function and $\varepsilon $ is random noise
with zero mean. If we estimate $f$ learning from a data set $\emph
L$ and obtain a model $f_{\emph{L}}$, the (quadratic)
generalization error on a test point $(t,\mathbf{x}) $ averaged
over all possible realizations of $\emph L$ (with respect to $P$
and noise $\varepsilon$) can be decomposed as:

\begin{eqnarray}
\mathrm{E}[(t-f_{{\emph L}}(\mathbf{x}))^{2}|{\emph L}
]&=&\mathrm{E}[\varepsilon ^{2}|\varepsilon] \\
&+&(\mathrm{E}[f_{\emph L}(\mathbf{x})|{\emph L}]-
f(\mathbf{x}))^{2} + \mathrm{E}[(f_{\emph L}
(\mathbf{x})-\mathrm{E}[f_{\emph L } (\mathbf{x})|{\emph
L}])^{2}|{\emph L}] \nonumber
\end{eqnarray}
The first term on the right-hand side is simply the noise variance
$\sigma _{\varepsilon }^{2}$; the second and third terms are,
respectively, the squared bias and variance of the estimation
method.

From the point of view of a single estimator $f_{\emph L}$, we can
interpret this equation by saying that a good method should be not
biased and have as little variance as possible between different
realizations. There are learning methods (for instance, ANNs) for
which the first condition is reasonably well met but the second
one is not satisfied since, for small changes or even with no
changes at all in ${\emph L}$, different learning experiments lead
to distinct predictors $f_{\emph L}$ (unstable learning methods).
A way to take advantage of this apparent weakness of these methods
is to make an aggregate of them.

If we rewrite the error decomposition in the form:

\begin{equation} \label{decomp}
\mathrm{E}[(t-\mathrm{E}[f_{\emph L}(\mathbf{x})|{\emph L}])^{2}
|{\emph L}]\equiv\mathrm{Bias}%
^{2}+\sigma _{\varepsilon }^{2} =
\mathrm{MeanError}-\mathrm{Variance},
\end{equation}
we can reinterpret this equation in the following way: using the
ensemble average $\Phi \equiv \mathrm{E}[f_{\emph L}|{\emph L}]$
as estimator, the generalization error can be reduced if we
produce fairly accurate models $f_{\emph L}$ (small
\textrm{MeanError}) that output diverse predictions for each test
point (large $\textrm{Variance}$). Of course, there is a trade-off
between these two conditions, but finding a good compromise
between the regressors' mean accuracy ($\equiv
1/\textrm{MeanError}$) and diversity ($\equiv \textrm{Variance}$)
seems particularly feasible for largely unstable methods like
ANNs.

For the results shown in Table 2a we have estimated separately the
accuracy and diversity components of the error according to
\ref{decomp}. In Table 5 we present the results obtained; for
easier comparison, we give them normalized by the mean accuracy
and diversity of the bagging ensemble members. As expected,
bagging produces the most accurate but less diverse predictors;
instead, the other aggregation methods resign some accuracy to
gain diversity. More interestingly, we see that, despite the
similar performance of SECA and SimAnn in Table 2a, these
aggregation methods select very different ensemble members. In
particular, SimAnn seeks mainly for diverse predictors although
they are not very accurate while SECA introduces diversity in a
more balanced way with accuracy. Epoch strategy is in general
intermediate between these two methods but not effective enough to
outperform them.


\section{Weighting Ensemble Members}

In the previous section we evaluated several ensemble construction
algorithms that essentially differ in the way they select the
particular stopping points for independently-trained ANNs. The
final aggregate prediction on a test point is simply the mean of
the individual predictions, without weighting the outputs of the
ensemble members ($w_n = 1/M$, $n=1,M$). This is not particularly
wise for SECA, since some of these members may have poor
generalization capabilities. SECA is a stepwise optimization
technique, and a known problem with these heuristics is that
during the optimization process they cannot review the choices
made in the past. Figure 1 shows a typical example of the problem
one can find for a given realization of the Friedman \#1 data set.
Open circles represent the evolution of training and test errors
during the construction of the ensemble using SECA. In this
example, the fourth added network clearly deteriorates the
ensemble performance, and this effect cannot be compensated by the
addition of more networks. Obviously, it also influences the
selection of the following ensemble members.\\

\begin{figure}[tbp]
\begin{center}
\includegraphics*[scale=1]{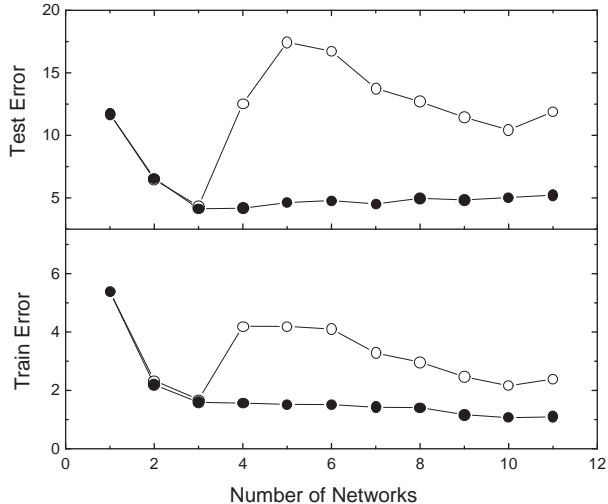}
\end{center}
\caption{Evolution of training and test errors during the ensemble
construction, in arbitrary units. Open circles correspond to SECA;
dots indicate the same evolution when ensemble members are
weighted according to W-SECA.} \label{Plot1}
\end{figure}

In a previous work \cite{Navone} we explored a possible way to
cope with this problem, using a slightly different SECA algorithm
that only accepts networks that improve the ensemble performance.
Unfortunately, new results showed that this algorithm also
produces some overfitting, being unable to clearly outperform
bagging on small and noisy data sets. A possible intermediate
solution is weighting the ensemble members, instead of rejecting
them if they do not improve the overall ensemble performance. This
allows us to reduce the influence of bad choices made in the past
by simply giving smaller weights to troublesome networks. Then,
following general ideas from boosting, we propose to modify the
algorithm so that the output of the ensemble at the $m$-th stage
becomes
\begin{equation}
\Phi_m ({\mathbf x}) = \sum_{n=1}^m w_n f_n ({\mathbf x}),
\end{equation}
where $w_n$ is a decreasing function of $e_n$, the MSE of the
$n$-th member over ${\emph D}$; {\it i.e.}, we weight each
ensemble member according to its individual performance on the
whole dataset. This is the way in which boosting reduces the
importance of overfitted members in the final ensemble. In
practice we have explored two different weighting functions:
\begin{equation}
w_i = \frac {e_i^{-\alpha}} {\sum_j e_j^{-\alpha}} , \ \ \  w_i =
\frac {\exp(-\alpha e_i)} {\sum_j \exp(-\alpha e_j)} .
\end{equation}
Figure 2 shows the results obtained with SECA on Friedman \#1 for
both weighting schemes. As we can see, for small to intermediate
values of $\alpha$ weighting produces better results than simply
averaging the individual predictions. For large values of $\alpha$
some overfitting is observed, since only a few particular networks
effectively contribute to the ensemble. As expected, this is more
acute for exponential weighting, but there are no other major
differences between both laws. On the other hand, the smaller the
noise the larger one can take $\alpha$ before overfitting is
observed. We have also considered Friedman \#2 and 3 data sets,
and the behavior in Figure 2 is
representative of the general trend.\\

\begin{figure}[tbp]
\begin{center}
\includegraphics*[scale=0.8]{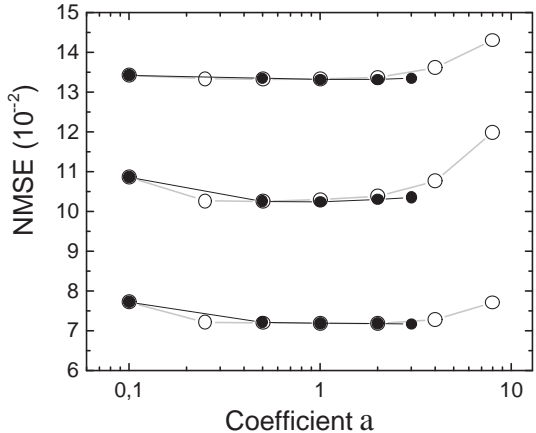}
\end{center}
\caption{Normalized mean-squared test error NMSE as a function of
the weighting coefficient $\alpha$ for Friedman \#1 data sets
(From top to bottom: high noise, low noise and noise-free data).
Open circles (dots) correspond to exponential (potential)
weighting.} \label{Plot2}
\end{figure}

In Figure 1 we have included the results of weighting SECA using
the power law with $\alpha = 2$ (this algorithm will be called
W-SECA) for the case discussed above. The problematic fourth
network is given a small weight, and is practically ignored by the
ensemble.

We have performed the $t$-test described before to establish
whether the performance obtained with W-SECA is significantly
better than that of SECA. To have a fair evaluation of the
algorithm just described we used the same ANNs considered in the
previous section. Tables 6a and 6b show the corresponding results,
which indicate that W-SECA outperforms SECA with statistical
significance in practically all situations studied.


We have also applied the weighting scheme to Bagging and SimAnn to
investigate if the effective elimination of some bad ensemble
members (by giving them small weights) has also impact on these
algorithms. One question to answer here is: Will this improvement
wash out the differences observed in Tables 1-4 between the
different algorithms? In order to have a fair evaluation of the
algorithms we used the same ANNs considered in the previous
section. The results obtained are collected in Tables 7a and 7b.


These tables show that for the 27 regression problems
corresponding to Friedman \#1, 2 and 3 data sets, W-SECA and
W-SimAnn outperform Bagging in 21 and 22 cases respectively.
Moreover, for the real-world databases and Ikeda map, W-SimAnn is
always better than W-Bagging, and W-SECA looses only on the Servo
database against W-Bagging. That is, although weighting is in
general beneficial for all algorithms, the member selection
strategy is still important to obtain good performances. This is
also supported by the results of paired $t$-tests between W-SECA
and W-SimAnn against W-Bagging (see Tables 8a and 8b).

It is also of interest to mention that the weighted algorithms
outperform non-weighted ones in 26 out of the 32 cases
investigated (compare best results in Tables 3 and 7). From the
remaining 6 cases, 4 correspond to high noise-scarce data
situations. Notice also that in these cases the best performers
are SECA and SimAnn. Finally, we stress that from the 32 problems
considered, W-Bagging performs better than Bagging in 27 cases,
W-SimAnn performs better than SimAnn in 29 cases, and W-SECA
performs better than SECA in 31 cases. We remark the important
improvements for SECA, which were expected according to the above
discussion in connection with Figure 1.


In Tables 7a and 7b we also present results obtained with the
boosting algorithms proposed in \cite{Druckerb} (``D-Boosting")
and \cite{Friedman} (``F-Boosting") for comparison. For these
algorithms we used a maximum of 20 boosting rounds, which should
produce a fair test considering the 20 ANNs ensembled in the
bagging-like methods. Notice that for the 27 Friedman datasets the
boosting algorithms perform better than W-SECA and W-SimAnn only
in three cases, and in these few cases the ``D-Boosting"
implementation is always the best performer. For the real-world
databases and Ikeda map this implementation and W-SimAnn are the
top performers.

As a final investigation on W-SECA and W-SimAnn, we have
considered the Mackey-Glass problem. This allows us to make a
comparison with seven other regression methods based on Support
Vector Machines (SVM) and regularized boosting using Radial Basis
Function (RBF) networks, as described in \cite{Muller} and
\cite{Ratsch}. Following these works, we introduced three levels
of uniform noise to the {\it training} set, with signal-to-noise
ratios of 6.2\%, 12.4\% and 18.6\% respectively, and Gaussian
noise with signal-to-noise ratios of 22.15\% and 44.30\%
respectively. The test set is kept noiseless to measure the true
prediction error. As mentioned in Section 4, to have a fair
comparison all the experimental settings (training and test set
lengths, embedding dimension, etc.) are the same as in
\cite{Muller} and \cite{Ratsch}. Table 9 presents the
corresponding results, which show that W-SECA and W-SimAnn are
among the top performers in most cases. We stress that they
perform worse than SVM methods only for the largest Gaussian noise
case (we are disregarding the CG-k result for the largest uniform
noise since it seems to be abnormally small).


\section{Summary and Conclusions}

We have performed a thorough evaluation of simple methods for the
construction of neural network ensembles. In particular, we
considered algorithms that can be implemented with an independent
(parallel) training of the ensemble members, and introduced a
framework that suggests naturally the SimAnn algorithm as the
optimal one. Taking as the ensemble prediction the simple average
of the ANN outputs, we have shown that SECA and SimAnn are the
best performers in the large majority of cases. These include
synthetic data with different noise levels and training set sizes,
and also real-world databases. We have also shown that these
methods resolve very differently the compromise between accuracy
and diversity through their particular search strategies.

The greedy method that we termed SECA seeks at every stage for a
new member that is at least partially anticorrelated with the
previous-stage ensemble estimator. This is achieved by applying a
late-stopping method in the learning phase of individual networks,
leading to a controlled level of overtraining of the ensemble
members. In principle this algorithm retains the simplicity of
independent network training, although, if necessary, it can avoid
the computational burden of saving intermediate networks in this
phase since it can be implemented in a sequential way. In this
implementation the method is a stepwise construction of the
ensemble, where each network is selected at a time and only its
parameters have to be saved. We showed, by comparison with several
other algorithms in the literature, that this strategy is
effective, as exemplified by the results in Tables 1 to 4.

The SimAnn algorithm, first proposed in this work, uses simulated
annealing to minimize the error on unseen data with respect to the
number of training epochs for each individual ensemble member.
This method is also very effective, being competitive with SECA on
most databases. Furthermore, the implementation of the
minimization step at the end of the ANNs training process is, in
practice, not very time consuming from a computational point of
view, being only a fraction of the time required to train the
networks.

We also discussed a known problem with stepwise selection
procedures like SECA, and proposed a modification of this
algorithm to overcome it. The modified algorithm, which we called
W-SECA, weights the predictions of ensemble members depending on
their individual performances. We showed that it improves the
results obtained with SECA in practically all cases. Moreover,
since weighting is in general beneficial for all the methods
considered, we investigated whether this procedure overrides the
differences between ensemble construction algorithms. We found
that the weighted versions of SECA and SimAnn (W-SECA and
W-SimAnn) are again the best performers, indicating the intrinsic
efficiency of these construction methods.

Finally, we have also performed a comparison of W-SECA and
W-SimAnn with several other regression methods, including methods
based on SVMs and regularized boosting. For this we used published
results in the literature corresponding to the Mackey-Glass
equation. Again in this case we found that the algorithms here
proposed are among the top performers in almost all situations
considered (Tables 7a,b and 9). Given this competitive behavior of
weighted bagging-like algorithms, one is tempted to speculate
that, for regression, the success of boosting ideas might not
be mainly related to the modification of resampling probabilities but
to the final error weighting of ensemble members.

We want to comment on the performance improvement obtained with
the aggregation algorithms discussed in this work. We found that
in general SECA and SimAnn, either in their weighted or
non-weighted versions, produce better results than other
algorithms in the literature (Bagging, NeuralBAG, Epoch). Although
this holds true in several cases with more than 95\% of
statistical significance, the performance improvement obtained
depends largely on the problem considered. For instance, with
respect to Bagging, the most common algorithm, one finds the
following (compare Tables 3 and 7): For the Friedman databases the
improvement can be very low with high noise (1\% or less), to very
large (up to 200\%) in some noise-free cases. For databases with
fixed noise level (real-world data and Ikeda map), the improvement
ranges from less than 1\% (Abalone) to nearly 12\% (Ikeda). The
answer to the question as to whether these performances justify
the use of the algorithms here proposed instead of Bagging would
depend, then, on the concrete application, particularly on how
critical it is. However, even for non-critical ones there is
always a chance that using W-SECA or W-SimAnn one might obtain
fairly large improvements. In any case, the best justification is
perhaps the fact that not much additional computational time is
required to implement these algorithms.

Before closing, we want to comment on a recent work\cite{Zhou}
partially related to the present one. In Ref. \cite{Zhou}, the
authors use genetic algorithms (GA) to select a suitable subset of
all the trained nets to build the ensemble. For this, they train a
number of ANNs to the optimal validation point like in Bagging,
and then assign to these networks an importance weight through a
GA strategy. Finally, only those networks that have weights larger
than a given threshold are kept in the ensemble. In the algorithm
--that they termed GASEN-- the predictions of the retained ANNs
are combined by simple average, which leads to good generalization
capabilities when compared to Bagging and Boosting. This strategy
can be readily implemented within our SimAnn algorithm by simply
allowing an appropriate random change in the number of aggregated
ANNs, in addition to the stochastic search of optimal training
epochs. Notice that this procedure would extend the GASEN
optimization to ANNs trained an arbitrary number of epochs
(instead of searching only among those at the optimal validation
point), which might be important since some degree of
single-network overtraining is known to improve the ensemble
performance. This combined approach, which would presumably bring
the best of SimAnn and GASEN into a single algorithm, is, however,
beyond the scope of this work.

In addition to the above proposal, as future work we are also
considering extending the methods here proposed to classification
problems and comparing their performances with those of boosting
strategies.


\section*{Acknowledgements} We acknowledge support for this
project from the National Agency for the Promotion of Science and
Technology (ANPCyT) of Argentina (grant PICT 11-11150).

\newpage

\vskip .2in
\begin{centering}
\begin{tabular}{c c c c c c c c}
\hline
 {\textbf {DB}} & {\textbf{Noise}} & {\textbf{Length}} & {\textbf{Single}} & {\textbf{Bagging}}
 & {\textbf{Epoch}} & {\textbf{SECA}} & {\textbf{SimAnn}} \\[-3pt]
 \hline
  &  & 50 & 4.48  & 3.59 & 3.72 & {\textbf{3.50}} & 3.53 \\[-9pt]
  & Free & 100 & 3.51 & 2.27 & 2.51 & {\textbf{2.13}} & 2.18 \\[-9pt]
  & & 200 & 1.10 & 0.49 & 0.48 & {\textbf{0.45}} & 0.46 \\ [-3pt]
 & & 50 & 5.59  &  4.69 &   4.87  &  4.66  &  {\textbf{4.63}} \\[-9pt]
 \#1&  Low & 100 & 4.43  &  3.18 &   3.46 & {\textbf{2.86}} &   2.89 \\[-9pt]
 &  & 200 & 2.71  &  1.92 &   2.12  & {\textbf{1.74}} &   1.80 \\[-3pt]
 &  & 50 & 7.47 &   6.33  &  6.54  &  6.34 & {\textbf{6.25}} \\[-9pt]
 & High & 100 & 5.97 &   4.98  &  5.16  &  4.80  &  {\textbf{4.78}} \\[-9pt]
 & & 200 & 4.59 &   3.62 &   3.78  &  3.39 &  {\textbf{3.38}} \\[-3pt] \hline
 &  & 20 & 2.07  &  1.38 &  1.62 & 1.40 & {\textbf{1.29}} \\[-9pt]
 & Free & 50 & 0.0177 & {\textbf{0.0102}} & 0.0104 &  0.0115 & 0.0119 \\[-9pt]
 &  &  100 & 0.0066 & {\textbf{0.0049}} & 0.0050 & 0.0053 & 0.0054 \\[-3pt]
 &  & 20 & 4.51  &  3.80 &   4.06 &   3.76 &  {\textbf{3.71}} \\[-9pt]
 \#2 & Low & 50 & 2.55 &  2.06 &  2.08  &  1.98 &  {\textbf{1.98}} \\[-9pt]
 &  & 100 & 1.78 & 1.61 &  {\textbf{1.60}} &   1.63  &  1.63 \\[-3pt]
 &  & 20 & 9.28  &  8.13  &  8.66 &   8.02 & {\textbf{7.88}}\\[-9pt]
 & High & 50 & 7.04 &   5.79 &  6.10 &   5.66  &  {\textbf{5.58}} \\[-9pt]
 &  & 100 & 5.60  &  5.10 & 5.15 & 5.08 & {\textbf{5.08}} \\[-3pt] \hline
 &   &  100 & 3.15 &   1.99 &  2.25 &   {\textbf{1.89}} & 2.04 \\[-9pt]
 & Free   & 200 & 1.49  &  0.95  &  1.00 & {\textbf{0.91}} & 0.94 \\[-9pt]
 &  & 400 & 0.65  &  0.48  &  0.49  &  {\textbf{0.47}} &  0.48 \\[-3pt]
 & & 100 & 8.77  &  7.10 &  7.42 &  {\textbf{6.80}} &   6.83 \\[-9pt]
 \#3 & Low  &  200 & 6.21  &  5.48  &  5.73 &   5.32 &  {\textbf{5.32}} \\[-9pt]
 & &  400 & 5.01 &   4.48  &  4.55 &   4.44 & {\textbf{4.44}} \\[-3pt]
 & &   100 & 19.35 &  {\textbf{13.35}} &  13.93 &  13.99 &  15.97 \\[-9pt]
 & High  &  200 & 13.96 &  {\textbf{11.48}} &  11.85 &  12.19 & 13.62 \\[-9pt]
 & &  400 &11.07  & 10.44 &  10.48 &  10.37  & {\textbf{10.34}} \\[-3pt] \hline
\end{tabular}
\end{centering}
\vskip .05in
{\textbf{Table 1a}}: Normalized mean-squared test errors (in units
of $10^{-2}$), averaged over 50 experiments, for Friedman \#1, 2
and 3 data sets. In this case 20\% of the data set $\emph D$ is
used for validation purposes. The results for Single correspond to
the average performance of a single ANN. The best result in each
case is highlighted in bold. \vskip .2in

\newpage

\vskip .2in
\begin{centering}
\begin{tabular}{l c c c c c}
  \hline
{\textbf{Database}} & {\textbf{Single}} & {\textbf{Bagging}} &
{\textbf{Epoch}}
& {\textbf{SECA}} &  {\textbf{SimAnn}} \\
\hline Abalone & 4.739 &  4.703 &  4.712 & {\textbf{4.686}} &
4.689
\\ Boston & 3.042 & 2.679 &  2.818 &  2.618 &  {\textbf{2.609}} \\
Ozone & 4.319 & 4.071 & 4.098 & 4.026 & {\textbf{4.024}}
\\ Servo & 2.578 & 2.194 & 2.232 & 2.209 & {\textbf{2.179}} \\
Ikeda &  28.73 & 19.17 & 19.43 & {\textbf{17.49}} & 17.64 \\
\hline
\end{tabular}
\end{centering}
\vskip .05in
{\textbf{Table 1b}}: Same as Table 1a for the databases indicated
in the first column. For Boston, Ozone and Servo (Abalone, Ikeda)
the results correspond to an average over 100 (50) independent
experiments.
\vskip .2in

\newpage

\vskip .2in
\begin{centering}
\begin{tabular}{c c c c c c c c}
\hline
 {\textbf{DB}}& {\textbf{Noise}} & {\textbf{Length}} & {\textbf{Single}} & {\textbf{Bagging}}
  & {\textbf{Epoch}} & {\textbf{SECA}} & {\textbf{SimAnn}} \\ [-3pt]
 \hline
  & & 50 & 5.30  & 3.57 & 3.74 & {\textbf{3.51}} & 3.54 \\ [-9pt]
  & Free & 100 & 4.35 & 2.49 & 2.54 & 2.32 & {\textbf{2.32}} \\ [-9pt]
  & & 200 & 1.91 & 1.05 & 1.07 & {\textbf{0.95}} & 0.96 \\ [-3pt]
  & & 50 & 6.12  &  4.60 &   4.83  &  4.54  &  {\textbf{4.54}} \\ [-9pt]
  \#1 & Low & 100 & 4.95  &  3.43 &  3.55 &  3.09 &   {\textbf{3.06}} \\ [-9pt]
   & & 200 & 3.29  &  2.13 &  2.28  & {\textbf{2.00}} &  2.04 \\ [-3pt]
   & & 50 & 8.04 &   6.38  &  6.61  &  6.31 & {\textbf{6.19}} \\ [-9pt]
   & High & 100 & 6.50 &  5.04  & 5.24  &  4.86  &  {\textbf{4.84}} \\ [-9pt]
   & & 200 & 5.20 &   3.92 &   3.88  & {\textbf{3.50}} & 3.54 \\ [-3pt] \hline
   & & 20 & 3.10  &  1.90 & 1.96 & 1.82 & {\textbf{1.72}} \\ [-9pt]
   & Free &     50 & 0.0346 & {\textbf{0.0144}} &  0.0146 & 0.0160 & 0.0156 \\ [-9pt]
   & &  100 & 0.0087 & {\textbf{0.0053}} & 0.0055 & 0.0056 & 0.0057 \\ [-3pt]
   & & 20 & 5.57  &  3.85 &  4.18 &   3.91 & {\textbf{3.82}} \\ [-9pt]
   \#2 & Low & 50 & 2.80 &   2.12 &  2.10  &  2.04  &  {\textbf{2.02}} \\ [-9pt]
   & & 100 & 2.05 &   {\textbf{1.62}} &   1.63  &  1.64 &   1.64 \\ [-3pt]
   & & 20 & 10.70  &  8.21  &  8.99 &   8.06 & {\textbf{7.85}}\\ [-9pt]
   & High & 50 & 7.19 &   5.87 &   6.25 &   5.71  &  {\textbf{5.65}} \\ [-9pt]
   & & 100 & 6.26  &  5.16 & 5.29 &  {\textbf{5.12}}  & 5.13 \\ [-3pt] \hline
  &   &  100 & 3.94 &  2.09 &  2.31  &  {\textbf{1.98}} & 2.04 \\ [-9pt]
 & Free   & 200 & 1.58  &  1.06  &  1.16 & {\textbf{1.02}} & 1.06 \\[-9pt]
 &   & 400 & 1.01  &  0.51  &  0.54  & 0.50 & {\textbf{0.50}} \\ [-3pt]
  & & 100 & 9.75  & 7.16 &  7.56 &   6.93 &   {\textbf{6.88}} \\ [-9pt]
  \#3 & Low  &  200 & 7.06  &  5.57  &  5.82 &   5.42 & {\textbf{5.39}} \\[-9pt]
 &  &  400 & 5.59 &   4.58  &  4.63 &   {\textbf{4.49}} & 4.50 \\ [-3pt]
  & &   100 & 20.62 &  {\textbf{13.47}} &  14.21 &  14.34 &  15.98 \\ [-9pt]
 & High  &  200 & 15.24 &  {\textbf{11.36}} &  12.02 &  11.86 & 12.99
\\ [-9pt]
  & &  400 & 11.84  & 10.56 &  10.58 &  10.42  & {\textbf{10.39}} \\ [-3pt] \hline
\end{tabular}
\end{centering}
\vskip .05in
{\textbf{Table 2a}}: Same as Table 1a but using 37\% of the
learning data for validation purposes.
\vskip .2in

\newpage

\vskip .2in
\begin{centering}
\begin{tabular}{l c c c c c}
\hline {\textbf{Database}} & {\textbf{Single}} &
{\textbf{Bagging}} &
{\textbf{Epoch}} & {\textbf{SECA}} &  {\textbf{SimAnn}} \\
\hline Abalone & 4.786 &  4.694 &  4.696 & {\textbf{4.669}} &
4.670
\\ Boston & 3.263 & 2.620 &  2.729 &  2.566 &  {\textbf{2.552}} \\
Ozone & 4.504 & 4.069 & 4.063 & 3.983 & {\textbf{3.973}}
\\ Servo & 3.403 & 2.340 & 2.348 & 2.299 & {\textbf{2.235}} \\ Ikeda
&  37.88 & 21.29 & 20.96 & 18.64 & \textbf{18.58} \\
\hline
\end{tabular}
\end{centering}
\vskip .05in
{\textbf{Table 2b}}: Same as Table 1b but using 37\% of the
learning data for validation purposes.
\vskip .2in

\newpage

\vskip .2in
\begin{centering}
\begin{tabular}{c c c c c c c c c}
\hline
 {\textbf{DB}} & {\textbf{Noise}} & {\textbf{Length}} & {\textbf{Single}} & {\textbf{Bagging}}
 & {\textbf{Epoch}} & {\textbf{NBAG}}  & {\textbf{SECA}} & {\textbf{SimAnn}} \\ [-3pt]
 \hline
  & & 50 & 4.53  & 3.21 & 3.43 & 3.31 & {\textbf{3.14}} & 3.15 \\ [-9pt]
  & Free & 100 & 3.39 & 1.93 & 1.95 & 1.92 & {\textbf{1.82}} & 1.82 \\ [-9pt]
  & & 200 & 0.91 & 0.33 & 0.31 & 0.31 & 0.30 & {\textbf{0.30}} \\ [-3pt]
  & & 50 & 5.22  &  4.17 &   4.36  &  4.21  &  4.15  &  {\textbf{4.12}} \\ [-9pt]
  \#1 & Low & 100 & 3.98  &  2.79 &   2.84 &   2.72 &   {\textbf{2.51}} &   2.53 \\ [-9pt]
  & & 200 & 2.62  &  1.66 &   1.68  &  1.66  &  {\textbf{1.50}} &   1.56 \\ [-3pt]
   & & 50 & 7.08 &   5.73  &  6.03  &  5.82 &   5.72  &  {\textbf{5.67}} \\ [-9pt]
   & High & 100 & 5.74 &   4.64  &  4.80  &  4.62  &  {\textbf{4.39}}  &  4.44 \\ [-9pt]
   & & 200 & 4.58 &   3.30 &   3.25  &  3.23 &  3.09  &  {\textbf{3.08}} \\ [-3pt] \hline
  & & 20 & 1.54  &  1.04 &   {\textbf{0.78}}  &  0.91  &  1.07 &
0.95 \\ [-9pt]
 &  Free &     50 & 0.0157 & 0.0083 & {\textbf{0.0081}} &  0.0084 & 0.0098 & 0.0088 \\ [-9pt]
  &  &  100 & 0.0059 & 0.0044 & {\textbf{0.0044}} & 0.0045 & 0.0047 & 0.0046 \\ [-3pt]
  & & 20 & 4.75  &  3.35 &   3.17 &   3.25 &   3.15 &   {\textbf{3.09}} \\ [-9pt]
  \#2 & Low & 50 & 2.68 &   1.89 &   1.86  &  1.84  &  1.88  &  {\textbf{1.84}} \\ [-9pt]
   & & 100 & 1.84 &   {\textbf{1.54}} &   1.55  &  1.54 &   1.56  &  1.56 \\ [-3pt]
   & & 20 & 9.97  &  7.61  &  8.31 &   7.82 &   7.51 &   {\textbf{7.35}}\\ [-9pt]
   & High & 50 & 7.12 &   5.65 &   5.68 &   5.67  &  {\textbf{5.54}}  &  5.55 \\ [-9pt]
   & & 100 & 5.65  &  4.91 &   {\textbf{4.89}}  &  4.90  &  4.90  &  4.93 \\ [-3pt] \hline
  &  &  100 & 2.79 &   1.64 &   1.91  &  1.73 &   {\textbf{1.60}} & 1.61 \\ [-9pt]
 & Free   & 200 & 1.05  &  0.73  &  0.83  &  0.74  &  0.72  &
{\textbf{0.71}} \\ [-9pt]
 &  & 400 & 0.61  &  0.39  &  0.39  &  {\textbf{0.38}} &  0.39  &  0.40 \\ [-3pt]
  & & 100 & 8.28  &  6.41 &   6.59 &   6.49 &   {\textbf{6.18}} &   6.19\\ [-9pt]
  \#3 & Low  &  200 & 6.30  &  5.10  &  5.19 &   5.11 &   4.94  &
{\textbf{4.92}} \\ [-9pt]
  & &  400 & 4.98 &   4.34  &  4.38 &   4.37 &   4.29  &  {\textbf{4.29}} \\ [-3pt]
 &  &  100 & 18.16 &  {\textbf{12.51}} &  13.04 &  12.97 &  13.45 &  14.80 \\ [-9pt]
 & High  &  200 & 14.58 &  {\textbf{11.09}} &  11.34 &  11.43 & 11.54
 & 12.46\\ [-9pt]
  & &  400 &11.13  & 10.15 &  10.18 &  10.14  & 10.09 &  {\textbf{10.07}} \\ [-3pt] \hline
\end{tabular}
\end{centering}
\vskip .05in
{\textbf{Table 3a}}: Normalized mean-squared test errors (in units
of $10^{-2}$), averaged over 50 experiments, for Friedman \#1, 2
and 3 data sets. In this case out-of-bag data are used for
validation purposes.
\vskip .2in

\newpage

\vskip .2in
\begin{centering}
\begin{tabular}{l c c c c c c}
  \hline
{\textbf{Database}} & {\textbf{Single}} & {\textbf{Bagging}} &
{\textbf{Epoch}}
& {\textbf{NBAG}} & {\textbf{SECA}} &  {\textbf{SimAnn}} \\
\hline Abalone & 4.728 &  4.644 &  4.634 &  4.649 &
{\textbf{4.629}} & 4.630
\\ Boston & 2.883 & 2.497 &  2.511 &  2.508 &  {\textbf{2.478}} &  2.495\\
Ozone & 4.245 & 3.931 & 3.975 & 3.921 &  3.893 &  {\textbf{3.873}}
\\ Servo & 2.668 & 1.930 & {\textbf{1.875}} & 1.900 & 1.891 & 1.905 \\ Ikeda
&  27.30 & 17.11 & 16.35 & 15.98 & {\textbf{15.22}} & 15.45\\
\hline
\end{tabular}
\end{centering}
\vskip .05in
{\textbf{Table 3b}}: Same as Table 3a for the databases indicated
in the first column. For Boston, Ozone and Servo (Abalone, Ikeda)
the results correspond to an average over 100 (50) independent
experiments.
\vskip .2in

\newpage

\vskip .2in
\begin{centering}
\begin{tabular}{l c c c c c c c c c}
  \hline
{\textbf{Friedman \#1}} & \multicolumn{3}{c}{{\textbf{Noise
Free}}}& \multicolumn{3}{c}{{\textbf{Low Noise}}}&
\multicolumn{3}{c}{{\textbf{High Noise}}} \\
Length & 50 & 100 & 200 & 50 & 100 & 200 & 50 & 100 & 200\\
SECA vs. Bag. & {\textbf{0.70}}& {\textbf{0.88}} &
{\textbf{0.88}} & {\textbf{0.66}}
& {\textbf{0.88}} & {\textbf{0.98}} & 0.54 & {\textbf{0.98}} & {\textbf{0.94}} \\
SimAnn vs. Bag. & {\textbf{0.66}} & {\textbf{0.76}} &
{\textbf{0.86}} & 0.60
& {\textbf{0.82}} & {\textbf{0.74}} & 0.56 & {\textbf{0.80}} & {\textbf{0.94}} \\
\hline {\textbf{Friedman \#2}} &
\multicolumn{3}{c}{{\textbf{Noise Free}}}&
\multicolumn{3}{c}{{\textbf{Low Noise}}}&
\multicolumn{3}{c}{{\textbf{High Noise}}} \\
Length & 20 & 50 & 100 & 20 & 50 & 100 & 20 & 50 & 100
\\
SECA vs. Bag. &   0.48 &   0.16  &  0.36 & {\textbf{0.74}} & 0.50
&   0.32  & 0.56 & {\textbf{0.66}} & 0.52
\\
SimAnn vs. Bag. & {\textbf{0.68}} &
0.28  &  0.14 &   {\textbf{0.80}} & 0.62 & 0.24 & {\textbf{0.74}} & {\textbf{0.66}}  &  0.40 \\
\hline
  {\textbf{Friedman \#3}} & \multicolumn{3}{c}{{\textbf{Noise
Free}}}& \multicolumn{3}{c}{{\textbf{Low Noise}}}&
\multicolumn{3}{c}{{\textbf{High Noise}}} \\
Length  & 100 & 200 & 400 & 100 & 200 & 400 & 100 & 200 &
400 \\
SECA vs. Bag. &   {\textbf{0.68}} &   0.58 &
0.48 & {\textbf{0.72}} &   {\textbf{0.82}} &  {\textbf{0.68}}  & 0.20 & 0.27 & {\textbf{0.66}} \\
SimAnn vs. Bag. & 0.56 & {\textbf{0.66}} & 0.44  &
{\textbf{0.70}} & {\textbf{0.82}} &
{\textbf{0.68}} & 0.06 & 0.08 & {\textbf{0.74}} \\
\hline
\end{tabular}
\end{centering}
\vskip .05in
{\textbf{Table 4a}}: Fraction of times SECA and SimAnn outperform
Bagging on Friedman databases in 50 independent experiments. Bold
numbers indicate results with a significance level above 95\%.
\vskip .2in

\vskip .2in
\begin{centering}
\begin{tabular}{l c c c c c}
  \hline
{\textbf{Database}} & {\textbf{Abalone}} & {\textbf{Boston}} &
{\textbf{Ozone}}
& {\textbf{Servo}} & {\textbf{Ikeda}} \\
\hline SECA vs. Bag. & {\textbf{0.74}}& 0.58 &   0.58 &   {\textbf{0.67}} &   {\textbf{1.00}} \\
\hline SimAnn vs. Bag. & 0.62 & 0.54 & {\textbf{0.68}} & 0.57 & {\textbf{0.96}} \\
\hline
\end{tabular}
\end{centering}
\vskip .05in
{\textbf{Table 4b}}: Same as Table 4a for the databases indicated
in the top row. For Boston, Ozone and Servo (Abalone, Ikeda) the
results correspond to an average over 100 (50) independent
experiments.

\newpage

\vskip .2in
\begin{centering}
\begin{tabular}{c c c c c c}
\hline
 {\textbf{DB}}& {\textbf{Noise}} & {\textbf{Length}}
  & {\textbf{Epoch}} & {\textbf{SECA}} & {\textbf{SimAnn}} \\ [-3pt]
 \hline
  & & 50 & 0.73 \  2.17 \ &  0.76 \   2.15 \ &   0.61 \   3.34 \\ [-9pt]
  & Free & 100 & 0.78 \  1.67   \  & 0.85 \  1.55  \  &   0.78 \  1.88  \\  [-9pt]
  & & 200 & 0.72 \  1.80   \  &  0.83 \  1.54   \  &  0.74 \  1.90  \\  [-3pt]
  & & 50 & 0.88 \ 1.39 \  &  0.85 \  1.75  \  &  0.81 \  2.01  \\  [-9pt]
   \#1 & Low & 100 & 0.68 \  2.75   \  &  0.75 \  2.60   \  &  0.62 \  3.72  \\  [-9pt]
   & & 200 &  0.69 \  2.06   \  &  0.70 \  2.26   \  & 0.56 \  3.18 \\  [-3pt]
   & & 50 &  0.88 \  1.47   \  &  0.87 \  1.69   \  & 0.82 \  2.11  \\  [-9pt]
   & High & 100 & 0.71 \  3.07  \  &   0.77 \  2.76  \  &   0.63 \  4.42  \\  [-9pt]
   & & 200 &  0.80 \  2.17   \  &  0.81 \  2.50   \  &  0.75 \  2.93 \\  [-3pt]  \hline
   & & 20 & 0.78 \  1.83   \  &  0.79 \  1.86  \  &  0.68 \  2.74  \\  [-9pt]
   & Free &     50  &  0.88 \  1.20   \  &  0.63 \  1.90   \  &  0.56 \  2.20  \\  [-9pt]
   & &  100 & 0.76 \  1.76   \  &  0.65 \  2.40   \  &  0.45 \  4.19 \\  [-3pt]
   & & 20 &  0.81 \  1.57   \  & 0.81 \  1.77  \  &   0.77 \  1.97  \\  [-9pt]
    \#2 & Low & 50 & 0.90 \  1.42   \  &  0.84 \  1.81  \  &  0.80 \  2.03  \\  [-9pt]
   & & 100 &   0.92 \  1.44   \  & 0.85 \  1.84  \  & 0.80 \  2.25 \\  [-3pt]
   & & 20  &  0.80 \  1.64   \  & 0.83 \  1.83   \  & 0.78 \  2.26  \\  [-9pt]
   & High & 50 &  0.84 \  1.64   \  &  0.83 \  2.11   \  & 0.78 \  2.53  \\  [-9pt]
   & & 100 & 0.92 \  1.44   \  & 0.88 \  1.99   \  &  0.84 \  2.33  \\  [-3pt]  \hline
  &   &  100 &  0.79 \  1.43   \  &  0.85 \  1.41  \  & 0.78 \  1.66  \\  [-9pt]
 & Free   & 200 &  0.78 \  1.51   \  &  0.85 \  1.50   \  &  0.79 \  1.64  \\ [-9pt]
 &   & 400 &  0.79 \  1.56   \  & 0.83 \  1.49  \  &  0.75 \  1.85  \\  [-3pt]
  & & 100 &  0.85 \  1.54   \  & 0.82 \  1.91   \  & 0.75 \  2.36  \\  [-9pt]
   \#3 & Low  &  200 &  0.86 \  1.61   \  & 0.81 \  2.20   \  &  0.75 \  2.78  \\ [-9pt]
 &  &  400 &  0.88 \  1.76   \  &  0.85 \  2.19  \  &  0.79 \  2.77  \\  [-3pt]
  & &   100 &  0.72 \  2.14  \  & 0.58 \  3.09  \  &  0.35 \  6.58  \\  [-9pt]
 & High  &  200 &  0.69 \  2.46   \  & 0.60 \  3.38   \  & 0.40 \  6.26  \\  [-9pt]
  & &  400 & 0.92 \  1.83   \  &  0.90 \  2.21   \  &  0.85 \  2.88  \\[-3pt] \hline
\end{tabular}
\end{centering}
\vskip .05in
{\textbf{Table 5}}: Accuracy and diversity of ensemble members for
different aggregation methods and Friedman data sets. Results are
normalized by the corresponding accuracy and diversity of the
bagging ensemble. \vskip .2in

\newpage
\vskip.2in
\begin{centering}

\begin{tabular}{l c c c c c c c c c}
  \hline
 {\textbf{Friedman \#1}} & \multicolumn{3}{c}{{\textbf{Noise
Free}}}& \multicolumn{3}{c}{{\textbf{Low Noise}}}&
\multicolumn{3}{c}{{\textbf{High Noise}}} \\
 Length   & 50 & 100 & 200 & 50 & 100 & 200 & 50 & 100 & 200 \\
 W-SECA vs. SECA  & {\textbf{0.64}} &   {\textbf{0.70}} &
 {\textbf{1.00}} &   0.60  &
    {\textbf{0.64}}  &  {\textbf{0.78}} &  {\textbf{0.64}}  &  0.40  &
 {\textbf{0.64}} \\ \hline
{\textbf{Friedman \#2}} & \multicolumn{3}{c}{{\textbf{Noise
Free}}}& \multicolumn{3}{c}{{\textbf{Low Noise}}}&
\multicolumn{3}{c}{{\textbf{High Noise}}} \\
 Length
   & 20 & 50 & 100 & 20 & 50 & 100 & 20 & 50 & 100 \\
W-SECA vs. SECA  & {\textbf{0.98}} &  {\textbf{0.86}} &
{\textbf{0.88}}  &  {\textbf{0.80}}
   &  {\textbf{0.78}} &  {\textbf{0.66}} &  {\textbf{0.64}} &   0.52  &
  0.34\\
  \hline
{\textbf{Friedman \#3}} & \multicolumn{3}{c}{{\textbf{Noise
Free}}}& \multicolumn{3}{c}{{\textbf{Low Noise}}}&
\multicolumn{3}{c}{{\textbf{High Noise}}} \\
 Length
    & 100 & 200 & 400 &100 & 200 & 400 & 100 & 200 & 400 \\
 W-SECA vs. SECA &  0.58 &   {\textbf{0.94}} &  {\textbf{0.78}}
 &   0.60 &   0.58  &  0.58 & 0.56 &   {\textbf{0.78}} &
  0.60\\ \hline
\end{tabular}
\end{centering}
\vskip .05in
{\textbf{Table 6a}}: Fraction of times W-SECA outperforms SECA on
Friedman databases in 50 independent experiments. Bold numbers
indicate results with a significance level above 95\%. \vskip .2in

\begin{centering}
\begin{tabular}{l c c c c c}
  \hline
{\textbf{Database}} & {\textbf{Abalone}} & {\textbf{Boston}} &
{\textbf{Ozone}}
& {\textbf{Servo}} & {\textbf{Ikeda}} \\
\hline W-SECA vs. SECA &   {\textbf{0.72}} &   0.54 &   0.56 &   {\textbf{0.60}} &  0.56 \\
\hline
\end{tabular}
\end{centering}
\vskip .05in
{\textbf{Table 6b}}: Same as Table 5a for the databases indicated.
\vskip .2in

\newpage

\vskip .2in
\begin{centering}
\begin{tabular}{l c c c c c c c c c}
  \hline
 {\textbf{Friedman \#1}} & \multicolumn{3}{c}{{\textbf{Noise
Free}}}& \multicolumn{3}{c}{\textbf{Low Noise}} &
\multicolumn{3}{c}{{\textbf{High Noise}}} \\
Length
    & 50 & 100 & 200 & 50 & 100 & 200 & 50 & 100 & 200 \\
W-Bagging &  3.23  &  1.88 &   0.13  &  4.17 &   2.75 &   1.62 &
5.73 & 4.63 &   3.27 \\
W-SECA & {\textbf{3.13}} &
{\textbf{1.76}} & 0.12 & {\textbf{4.10}} & {\textbf{2.49}} &
{\textbf{1.47}} & {\textbf{5.69}} & {\textbf{4.40}} & 3.07
\\
W-SimAnn & 3.24
& 1.77 & {\textbf{0.11}} & 4.13 & 2.54 & 1.50 & 5.73  &  4.44  & {\textbf{3.07}}\\
F-Boosting & 3.63 & 2.46 & 0.53 & 4.46 & 3.24 & 2.15 & 6.13 & 4.90
&
4.00\\
D-Boosting & 3.32 & 1.98 & 0.60 & 4.10 & 2.68 & 1.67 & 5.78 & 4.50
&
3.21 \\
\hline
 {\textbf{Friedman \#2}} & \multicolumn{3}{c}{{\textbf{Noise
Free}}}& \multicolumn{3}{c}{{\textbf{Low Noise}}}&
\multicolumn{3}{c}{{\textbf{High Noise}}} \\
Length
   & 20 & 50 & 100 & 20 & 50 & 100 & 20 & 50 & 100 \\
W-Bagging &  0.65  &  0.0076 & 0.0041 & 3.28  &  1.86  &
{\textbf{1.53}} & 7.52 & 5.64 & {\textbf{4.90}} \\
W-SECA &
0.62  &
0.0079 &  0.0042 & 2.94 & 1.84 & 1.55 & 7.50 & {\textbf{5.56}} & 4.91\\
W-SimAnn & {\textbf{0.49}} & {\textbf{0.0076}} &
{\textbf{0.0041}} & {\textbf{2.93}} & 1.82
& 1.57 & 7.50 & 5.59 & 4.94 \\
F-Boosting & 0.83 & 0.0116 & 0.0050 & 3.09 & 1.98 & 1.64 & 7.74 &
5.80 & 5.14 \\
D-Boosting & 1.20 & 0.0081 & 0.0044 & 3.14 & {\textbf{1.81}} &
1.56 & {\textbf{7.49}} &
5.57 & 4.96 \\
\hline
 {\textbf{Friedman \#3}} & \multicolumn{3}{c}{{\textbf{Noise
Free}}}& \multicolumn{3}{c}{{\textbf{Low Noise}}}&
\multicolumn{3}{c}{{\textbf{High Noise}}} \\
Length
    & 100 & 200 & 400 &100 & 200 & 400 & 100 & 200 & 400 \\
W-Bagging  & 1.71  &  0.67 &   0.37 &   6.42 &   5.09 &   4.36 &
{\textbf{12.50}} & {\textbf{11.08}}  & 10.14 \\
W-SECA &
{\textbf{1.64}}
&   0.66 & 0.36 & 6.16 & 4.93 & 4.28 & 13.38 &  11.39 & 10.08\\
W-SimAnn & 1.69 & {\textbf{0.65}} & {\textbf{0.35}} & 6.24
& {\textbf{4.90}} & {\textbf{4.26}}
& 14.83 & 12.14 & {\textbf{10.07}}\\
F-Boosting & 2.37 & 0.86 & 0.55 & 7.47 & 5.80 & 4.75 & 17.16 &
13.37
& 10.74 \\
D-Boosting & 1.77 & 0.72 & 0.40 & {\textbf{6.15}} & 5.00 & 4.28 &
13.24 & 11.52 & 10.10 \\ \hline
\end{tabular}
\end{centering}
\vskip .05in
{\textbf{Table 7a}}: Normalized mean-squared test errors (in units
of $10^{-2}$) for the weighted versions of the algorithms
indicated. These figures correspond to an average over 50
experiments, using out-of-bag data for validation purposes. The
results of two different boosting algorithms are also included for
comparison. \vskip .2in

\newpage

\begin{centering}
\begin{tabular}{l c c c c c}
  \hline
{\textbf{Database}} & {\textbf{Abalone}} & {\textbf{Boston}} &
{\textbf{Ozone}}
& {\textbf{Servo}} & {\textbf{Ikeda}} \\
\hline
W-Bagging & 4.644 & 2.503 & 3.931 & 1.840 & 16.64 \\
W-SECA & 4.626 & 2.482 & 3.887 & 1.845 & 15.10 \\
W-SimAnn & 4.631 & 2.498 & {\textbf{3.865}} &
 1.823 & {\textbf{15.10}} \\
 F-Boosting & 4.646 & 2.638 & 4.028 & 2.172 & 22.07 \\
 D-Boosting & {\textbf{4.624}} & {\textbf{2.479}} & 3.920 & {\textbf{1.778}} & 16.19 \\
\hline
\end{tabular}
\end{centering}
\vskip .05in
{\textbf{Table 7b}}: Same as Table 6a for the real-world databases
indicated. \vskip .2in

\vskip .2in
\begin{centering}
\begin{tabular}{l c c c c c c c c c}
  \hline
{\textbf{Friedman \#1}} & \multicolumn{3}{c}{{\textbf{Noise
Free}}}& \multicolumn{3}{c}{{\textbf{Low Noise}}}&
\multicolumn{3}{c}{{\textbf{High Noise}}} \\
Length & 50 & 100 & 200 & 50 & 100 & 200 & 50 & 100 & 200
\\
W-SECA vs. W-Bag. & {\textbf{0.68}}  &  {\textbf{0.86}} &
{\textbf{0.64}}  & {\textbf{0.68}} & {\textbf{0.78}} &
{\textbf{0.92}}
 &  0.58 & {\textbf{0.84}} & {\textbf{0.84}} \\
W-SimAnn vs. W-Bag. & 0.46 & {\textbf{0.78}} &
{\textbf{0.96}} & 0.58 & {\textbf{0.66}} & {\textbf{0.74}} & 0.40
& {\textbf{0.82}} & {\textbf{0.88}}
\\
  \hline
{\textbf{Friedman \#2}} & \multicolumn{3}{c}{{\textbf{Noise
Free}}}& \multicolumn{3}{c}{{\textbf{Low Noise}}}&
\multicolumn{3}{c}{{\textbf{High Noise}}} \\
Length
   & 20 & 50 & 100 & 20 & 50 & 100 & 20 & 50 & 100 \\
W-SECA vs. W-Bag. & 0.60 &   0.38 &   0.44  &  {\textbf{0.86}}  &
0.56 & 0.32  & 0.56 & 0.62 & 0.52 \\
W-SimAnn vs. W-Bag. &
{\textbf{0.92}} & 0.54  & 0.62 & {\textbf{0.88}} & 0.54 & 0.16 &
0.56 & 0.48 & 0.38\\ \hline
{\textbf{Friedman \#3}} &
\multicolumn{3}{c}{{\textbf{Noise Free}}}&
\multicolumn{3}{c}{{\textbf{Low Noise}}}&
\multicolumn{3}{c}{{\textbf{High Noise}}} \\
Length
    & 100 & 200 & 400 &100 & 200 & 400 & 100 & 200 & 400 \\
W-SECA vs. W-Bag. & 0.60 &   0.60 &   0.60  &  {\textbf{0.74}} &
{\textbf{0.90}} & {\textbf{0.82}} & 0.18 & 0.22 & {\textbf{0.70}}
\\
W-SimAnn vs. W-Bag. &
0.58 & {\textbf{0.72}} & {\textbf{0.66}} & {\textbf{0.70}} &
{\textbf{0.86}}
 & {\textbf{0.78}} & 0.10 & 0.06 & {\textbf{0.64}} \\
\hline
\end{tabular}
\end{centering}
\vskip .05in
{\textbf{Table 8a}}: Same as Table 4a for the weighted versions of
the algorithms indicated. \vskip .2in

\begin{centering}
\begin{tabular}{l c c c c c}
  \hline
{\textbf{Database}} & {\textbf{Abalone}} & {\textbf{Boston}} &
{\textbf{Ozone}} & {\textbf{Servo}} & {\textbf{Ikeda}} \\ \hline
W-SECA vs. W-Bag. & {\textbf{0.80}} & {\textbf{0.63}} &
{\textbf{0.61}}& {\textbf{0.62}} &   {\textbf{0.98}} \\

W-SimAnn vs. W-Bag. & {\textbf{0.68}} & 0.57 &
{\textbf{0.65}} & 0.55 & {\textbf{0.94}}\\
\hline
\end{tabular}
\end{centering}
\vskip .05in
{\textbf{Table 8b}}: Same as Table 4b for the weighted versions of
the algorithms indicated.
\newpage

\vskip .2in
\begin{centering}
\begin{tabular}{c c c c c c}
  \hline
 {\textbf{Mackey-Glass}} & \multicolumn{3}{c}{{\textbf{Uniform Noise}}}
 & \multicolumn{2}{c}{\textbf{Gaussian Noise}}\\
Noise Level
    & 6.20\% & 12.40\% & 18.60\% & 22.15\% & 44.30\% \\ \hline
CG-k &  0.11  &  0.35 &   0.31  &  - &  - \\
CG-ak & 0.10 & 0.35 & 0.65 & - & - \\
BAR-k & 0.13 & 0.32 & 0.51 & - & - \\
BAR-ak & 0.12 & 0.27 & 0.66 & - & - \\
SVM e-ins & 0.07 & 0.28 & 0.57 & 0.58 & 3.23 \\
SVM Huber & 0.13 & 0.38 & 0.71 & 0.58 & 3.23 \\
RBF-NN &  0.16  & 0.38 & 1.54 & 0.65  &  3.90 \\ \hline W-Bagging
& 0.07 & 0.25 &  0.58 & 0.69 & 4.00 \\
W-SECA & 0.07 & 0.25 & 0.53 & 0.66 & 3.67 \\
W-SimAnn & 0.08 & 0.24 & 0.55 & 0.57 & 3.78 \\ \hline
\end{tabular}
\end{centering}
\vskip .05in
{\textbf{Table 9}}: Test set prediction errors (in units of
$10^{-2}$) for the Mackey-Glass problem using W-SECA and W-SimAnn.
For comparison, we give the results of other methods in the
literature taken from \cite{Ratsch}.

\end{document}